\documentclass[10pt, conference]{IEEEtran}
\IEEEoverridecommandlockouts

\usepackage[utf8]{inputenc}
\usepackage{cite}

\usepackage{amsmath,amssymb,amsfonts}
\usepackage{bbm} 

\usepackage{algorithm}
\usepackage{algpseudocode}

\usepackage{graphicx}
\usepackage{booktabs} 
\usepackage{textcomp}
\usepackage{xcolor}

\usepackage{url}
\graphicspath{{figures/}}

\begin{document}

\title{Lightweight Intrusion Detection in IoT via SHAP-Guided Feature Pruning and Knowledge-Distilled Kronecker Networks
\thanks{This work has been published in the proceedings of the 2025 8th International Conference on Advanced Communication Technologies and Networking (CommNet).}}

\author{\IEEEauthorblockN{Hafsa BENADDI\IEEEauthorrefmark{1}, Mohammed JOUHARI\IEEEauthorrefmark{4}, Nouha LAAMECH\IEEEauthorrefmark{5}, Anas MOTII\IEEEauthorrefmark{5}, Khalil IBRAHIMI\IEEEauthorrefmark{4}}
\IEEEauthorblockA{\IEEEauthorrefmark{1} LSIA Laboratory, Moroccan School of Engineering Sciences (EMSI), Tanger, Morocco \\
\IEEEauthorrefmark{4} Laboratory of Research in Informatics (LaRI), Faculty of Sciences, Ibn Tofail University, Kenitra, Morocco\\
\IEEEauthorrefmark{5} College of Computing, University Mohammed VI Polytechnic, Benguerir, Morocco}
Emails: h.benaddi@emsi.ma, \{mohammed.jouhari1, khalil.ibrahimi\}@uit.ac.ma, \{nouha.laamech, anas.motii\}@um6p.ma}

\maketitle

\IEEEpubid{
\begin{minipage}{\columnwidth}
\raggedright
979-8-3315-5781-2/25/\$31.00 ©2025 IEEE
\end{minipage}
\hspace{\columnsep}\makebox[\columnwidth]{}%
}
\IEEEpubidadjcol

\begin{abstract}
The widespread deployment of Internet of Things (IoT) devices requires intrusion detection systems (IDS) with high accuracy while operating under strict resource constraints. Conventional deep learning IDS are often too large and computationally intensive for edge deployment. We propose a lightweight IDS that combines SHAP-guided feature pruning with knowledge-distilled Kronecker networks. A high-capacity teacher model identifies the most relevant features through SHAP explanations, and a compressed student leverages Kronecker-structured layers to minimize parameters while preserving discriminative inputs. Knowledge distillation transfers softened decision boundaries from teacher to student, improving generalization under compression. Experiments on the \textsc{TON\_IoT} dataset show that the student is nearly three orders of magnitude smaller than the teacher yet sustains macro-F1 above 0.986 with millisecond-level inference latency. The results demonstrate that explainability-driven pruning and structured compression can jointly enable scalable, low-latency, and energy-efficient IDS for heterogeneous IoT environments.
\end{abstract}


\section{Introduction}

The rapid expansion of the Internet of Things (IoT) has transformed industrial, domestic, and public domains through ubiquitous sensing and automation. This growth, however, introduces new security vulnerabilities, as low-power IoT devices typically operate under severe constraints on processing, memory, and energy, and are often deployed in unattended environments. Conventional security mechanisms relying on complex models and frequent updates are unsuitable for such conditions. Intrusion detection systems (IDS) remain a critical line of defence, yet their deployment in IoT networks is restricted by requirements for scalability across heterogeneous devices, real-time responsiveness, and minimal energy consumption without compromising accuracy \cite{11161305}.

Recent studies highlight that traditional and deep learning IDS, when shifted from centralized high-performance servers to edge devices, suffer from excessive memory and computational demands. Such models cause high inference delays and energy costs, making them impractical for battery-powered nodes. This has motivated research into lightweight IDS architectures capable of balancing efficiency and accuracy \cite{BENADDI2025101624}. Approaches based on model compression, including quantisation and knowledge distillation, demonstrate that significant parameter reduction can be achieved while sustaining competitive accuracy. Similarly, ensemble methods have improved robustness and scalability in complex traffic scenarios \cite{akif2025hybrid}. However, existing evaluations infrequently report deployment-oriented metrics such as CPU load or device-level energy consumption \cite{10658099}, leaving the interaction among latency, scalability, and detection performance not sufficiently understood.

This paper addresses these challenges by proposing a latency-sensitive and energy-efficient IDS framework tailored to heterogeneous IoT environments. The design employs a teacher–student paradigm in which a high-capacity teacher model computes Shapley additive (SHAP) feature importances to guide feature pruning, while a Kronecker-compressed student network is trained via knowledge distillation to mimic teacher decision boundaries with drastically fewer parameters. By aligning model complexity with resource constraints, the framework explicitly balances latency, accuracy, and energy consumption. The system is evaluated on the public \textsc{TON\_IoT} dataset, and experiments on representative hardware confirm that the proposed student achieves substantial reductions in model size and inference latency with minimal accuracy degradation.To support reproducibility and comparative research, the full implementation and dataset splits are released as open source.

The main contributions are as follows:
\begin{itemize}
  \item Identification of deployment challenges in IoT IDS, with emphasis on trade-offs among scalability, latency, energy efficiency, and accuracy \cite{BENADDI2025101624}.
  \item A hierarchical teacher–student IDS framework integrating SHAP-guided feature pruning with Kronecker-compressed layers for resource-adaptive operation.
  \item Comprehensive evaluation on the \textsc{TON\_IoT} dataset and edge hardware, demonstrating that the student achieves drastic compression and low-latency inference while preserving high detection accuracy.
  \item An open-source implementation and benchmark dataset splits to facilitate reproducibility and future comparative studies\footnote{\url{https://github.com/mohammed-jouhari/iot-ids-lightweight}}.
\end{itemize}

The remainder of the paper is organized as follows. Section~\ref{sec:related_work} reviews prior work on lightweight IDS, explainable pruning, knowledge distillation, and Kronecker factorization. Section~\ref{sec:proposed} introduces the proposed framework, Section~\ref{sec:evaluation} presents the experimental setup and results, Section~\ref{sec:discussion} discusses implications and research directions, and Section~\ref{sec:conclusion} concludes the paper.

\section{Related Work}
\label{sec:related_work}

Research on intrusion detection for the IoT has progressed across several complementary directions, including lightweight model design, explainable feature selection, knowledge distillation, and structured parameter reduction. Six representative contributions from leading journals are highlighted here to situate the present work.

Federated learning has been combined with explainable feature aggregation, where local clients share SHAP-derived importance profiles to a central server that constructs a global model with high detection accuracy while maintaining privacy constraints \cite{hossain2025federated}. Although effective, this approach omits structured compression. Meta-heuristic optimisation has also been employed to select features and hyperparameters, with teaching–learning based optimisation reducing communication overhead while sustaining detection rates against diverse attack types \cite{kaushik2023novel}. Such optimisation, however, remains confined to classical classifiers and does not support interpretability or model transfer.

Deep learning-based IoT IDS have been made more transparent by incorporating SHAP and LIME explanations. A one-dimensional convolutional detector achieves near-perfect F1-scores on the ToN-IoT dataset, while offering both global and local interpretability \cite{ebrahimi2025explainable}. Despite its transparency, the model lacks compression or distillation mechanisms. Structured compression has been explored through Kronecker product factorisation, where KronNet reduces parameter counts by orders of magnitude and still delivers accuracy above 99\% on benchmark datasets \cite{ullah2025kronnet}. This demonstrates the promise of Kronecker-based designs, but without feature-guided pruning or teacher–student transfer.

Knowledge distillation surveys underscore its potential for deploying compact models on constrained devices, classifying techniques by knowledge type, training strategy, and application \cite{gou2021knowledge}. However, most studies focus on the vision and language domains rather than intrusion detection. Finally, adaptive federated IDS frameworks have integrated Long Short-Term Memory with joint strategy optimisation, dynamically adjusting hyperparameters across heterogeneous clients while maintaining privacy \cite{alsorour2025lstmjso}. These systems achieve robustness under non-stationary traffic, but do not incorporate explainable feature pruning or structural compression.

The reviewed literature collectively advances federated detection, meta-heuristic optimization, explainability, Kronecker compression, and distillation, but each dimension has evolved largely in isolation. The present work addresses this fragmentation by unifying SHAP-guided pruning with knowledge-distilled Kronecker networks, resulting in an interpretable and compact IDS suitable for heterogeneous IoT deployments.

\section{Proposed Model}
\label{sec:proposed}

\begin{figure}[t]
  \centering
  \includegraphics[width=0.95\linewidth]{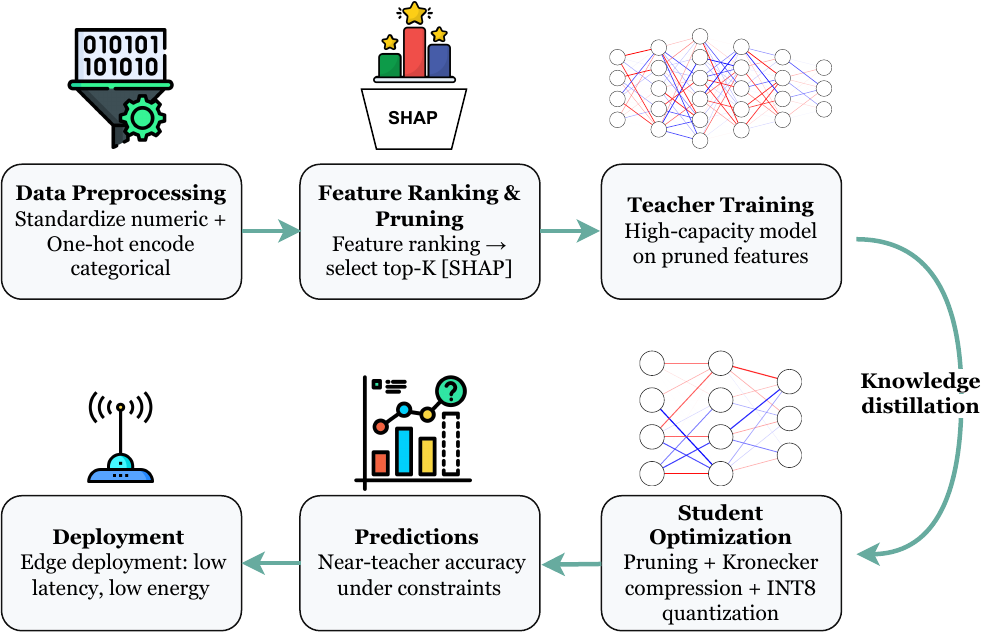}
  \caption{Architectural overview of the proposed IDS: data preprocessing, SHAP-based feature ranking and pruning, teacher training, Kronecker-compressed student optimization with knowledge distillation, and final deployment on resource-constrained IoT devices.}
  \label{fig:proposed_model}
\end{figure}

This work aims to design an intrusion detection system (IDS) that preserves high detection accuracy while remaining feasible for deployment on resource-constrained IoT devices. To achieve this objective, we adopt a teacher–student framework in which a high-capacity teacher model is trained on the complete feature space and subsequently guides a compressed student model. The proposed framework is depicted at two levels: Fig.~\ref{fig:proposed_model} provides a schematic overview of the system architecture, while Algorithm~\ref{alg:pipeline} details the corresponding training procedure.

\subsection{Preprocessing and Teacher Network}
As shown in the first block of Fig.~\ref{fig:proposed_model}, raw traffic flows undergo preprocessing where numeric features are standardized and categorical ones are one-hot encoded. This corresponds to line~1 in Algorithm~\ref{alg:pipeline}. Each flow is represented as $\mathbf{x}\in\mathbb{R}^{d}$ with ground-truth label $y\in\{0,1\}$. After preprocessing, the expanded input vector $\widetilde{\mathbf{x}}\in\mathbb{R}^{p}$ has $p=1{,}624$ dimensions.  

Lines~2–3 in Algorithm~\ref{alg:pipeline} correspond to the training of the teacher model. The teacher is a multilayer perceptron (MLP) with hidden layers of widths $512$, $256$, and $128$, each employing batch normalization, ReLU activation, and dropout. Given $\widetilde{\mathbf{x}}$, the teacher outputs logits $\mathbf{z}=f(\widetilde{\mathbf{x}})$ and class probabilities $\mathbf{p}=\mathrm{softmax}(\mathbf{z})$. The training minimizes weighted cross-entropy:  
\begin{equation}
  \mathcal{L}_{\text{CE}} = -\sum_{c=1}^{C} w_{c}\,\mathbbm{1}[y=c] \,\log p_{c},
  \label{eq:ce_loss}
\end{equation}
where $w_{c}$ denotes the inverse class frequency. The teacher is optimized with AdamW (learning rate $10^{-3}$, weight decay $10^{-3}$), cosine annealing scheduling, and early stopping with patience of eight epochs. Although focal loss was considered, weighted cross-entropy consistently produced the best results.

\subsection{Feature Ranking and Pruning}
The “SHAP Feature Ranking \& Pruning” module in Fig.~\ref{fig:proposed_model} corresponds to lines~3–4 in Algorithm~\ref{alg:pipeline}. SHAP values provide attribution scores per feature, averaged across samples to form a global importance vector $\mathbf{s}\in\mathbb{R}^{p}$. Features are ranked accordingly, and the top-$K$ subset $\pi_{1:K}$ is retained. An ablation analysis with $K\in\{32,64,96,128\}$ showed that $K=32$ preserved macro-F1 within 2\% of the full model, making it the adopted configuration for the student.

\subsection{Kronecker-Compressed Student Network}
Fig.~\ref{fig:proposed_model} highlights “Student Optimization,” which corresponds to line~5 in Algorithm~\ref{alg:pipeline}. The student is constructed with Kronecker factorized layers. Each linear transformation is expressed as  
\begin{equation}
  \mathbf{y} = \mathbf{x}\,\mathbf{W}^{\top} + \mathbf{b}, \quad \mathbf{W}=\mathbf{A}\otimes\mathbf{B},
  \label{eq:kron_layer}
\end{equation}
where $\mathbf{A}\in\mathbb{R}^{a_{1}\times a_{2}}$ and $\mathbf{B}\in\mathbb{R}^{b_{1}\times b_{2}}$ with $a_{1}b_{1}=m$ and $a_{2}b_{2}=n$. The student processes the top-$K$ features, pads to a compatible dimension, and applies two Kronecker layers separated by batch normalization and ReLU, followed by a linear head. This design reduces the parameter count to $1{,}282$, compared with $998{,}274$ in the teacher, making the student model highly resource-efficient.

\subsection{Knowledge Distillation}
The “Knowledge Distillation” component in Fig.~\ref{fig:proposed_model} is realized in lines~6–9 of Algorithm~\ref{alg:pipeline}. The student learns not only from hard labels but also from the teacher’s softened predictions. Let $\mathbf{z}_{s}$ and $\mathbf{z}_{t}$ denote the logits of the student and teacher, respectively. The distillation loss is:  
\begin{equation}
  \mathcal{L}_{\text{KD}} = T^{2}\,\mathrm{KL}\bigl(\sigma(\mathbf{z}_{t}/T)\,\Vert\,\sigma(\mathbf{z}_{s}/T)\bigr),
  \label{eq:kd_loss}
\end{equation}
where $\sigma$ is the softmax function and $T$ is the temperature. The total loss combines both sources of supervision:  
\begin{equation}
  \mathcal{L}_{\mathrm{tot}}=(1-\alpha)\mathcal{L}_{\text{CE}}+\alpha\mathcal{L}_{\text{KD}},
\end{equation}
with $\alpha\in[0,1]$. A grid search over $T\in\{2,3,4\}$ and $\alpha\in\{0.5,0.7,0.9\}$ identified the most effective settings by maximizing validation macro-F1.

\subsection{Deployment Considerations}
The final block of Fig.~\ref{fig:proposed_model} emphasizes deployment, corresponding to line~10 in Algorithm~\ref{alg:pipeline}. After training and distillation, the student is optionally quantized to INT8 precision, further reducing memory and computational cost. With only $1{,}282$ parameters, the student model is lightweight enough for microcontrollers and edge gateways. This compact design ensures low-latency inference and reduced energy consumption, while maintaining detection accuracy comparable to the teacher. These characteristics directly address the constraints of real-world IoT scenarios where continuous monitoring must operate under strict device limitations.

\begin{algorithm}[t]
  \caption{Training workflow of the proposed IDS framework}
  \label{alg:pipeline}
  \begin{algorithmic}[1]
    \Require $(\mathbf{X}_{\text{train}},\mathbf{y}_{\text{train}})$, $(\mathbf{X}_{\text{val}},\mathbf{y}_{\text{val}})$, $(\mathbf{X}_{\text{test}},\mathbf{y}_{\text{test}})$
    \State Preprocess numeric and categorical features.
    \State Train teacher with Eq.~\eqref{eq:ce_loss}.
    \State Compute SHAP attributions and rank features.
    \State Select smallest $K$ preserving macro-F1.
    \State Build student with Kronecker layers on top-$K$ features.
    \For{$T\in\{2,3,4\}$, $\alpha\in\{0.5,0.7,0.9\}$}
      \State Train student with $(1-\alpha)\mathcal{L}_{\text{CE}}+\alpha\mathcal{L}_{\text{KD}}$.
      \State Validate macro-F1 and retain best $(T,\alpha)$.
    \EndFor
    \State Optionally quantize and evaluate fp32/int8 variants.
  \end{algorithmic}
\end{algorithm}

\noindent
In summary, the framework follows a coherent pipeline: raw traffic is standardized and encoded, a high-capacity teacher is trained, SHAP-based ranking identifies the most informative attributes, and a Kronecker-compressed student is distilled to approximate the teacher’s decision boundary. The integration of pruning, structured compression, and knowledge distillation ensures that the resulting model achieves near-teacher accuracy while remaining lightweight enough for on-device inference. When coupled with post-training quantization, this design results in an IDS that balances detection reliability with latency and energy efficiency, making it particularly suited for deployment in resource-constrained IoT environments.  

Having established the design of the teacher–student IDS and its deployment-oriented optimizations, we now turn to the empirical assessment of its effectiveness. The following section details the experimental setup, including dataset partitioning, training protocols, and hyperparameter configurations, and evaluates the proposed framework in terms of detection accuracy, latency, throughput, and memory efficiency. This evaluation provides a quantitative basis for judging whether the architectural choices outlined above translate into tangible benefits under realistic IoT constraints.

\section{Evaluation}
\label{sec:evaluation}

This section evaluates the proposed SHAP-guided pruning and Kronecker-based distillation framework under realistic IoT intrusion detection scenarios. The analysis focuses on both algorithmic accuracy and deployment-oriented metrics, with particular attention to runtime efficiency, parameter reduction, and robustness under imbalanced traffic distributions. We first describe the training setup and dataset profiling before reporting experimental results.

\subsection{Model Training Setup}
\label{sec:exp_setup}

All experiments were conducted in a Google Colab environment with GPU acceleration and a \texttt{PyTorch} runtime. The \textsc{TON\_IoT} dataset was preprocessed using the same ColumnTransformer pipeline described in Section~\ref{sec:dataset_char}, where numerical features were standardized and categorical attributes one-hot encoded. To avoid source leakage, a source-aware split was applied whenever device or subnet identifiers were available; otherwise stratified random partitioning preserved class ratios. The final distribution allocated 70\% of data to training, 15\% to validation, and 15\% to testing, ensuring that evaluation measured cross-source generalization rather than memorization.

The hyperparameters are summarized in Table~\ref{tab:hyperparams}. Both teacher and student models were trained with a batch size of 1024 using AdamW (learning rate $10^{-3}$, weight decay $10^{-3}$). A cosine learning rate scheduler with step size 10 and decay factor $\gamma=0.5$ stabilized convergence under non-IID traffic. Dropout with rate 0.3 was applied, and early stopping with patience of 8 epochs prevented overfitting. Weighted cross-entropy was adopted as the primary loss. SHAP-guided feature pruning was always applied before Kronecker-based knowledge distillation to ensure reproducibility. Exported artifacts include preprocessing maps, class distributions, and convergence curves for auditability.

\begin{table}[t]
  \centering
  \caption{Hyperparameters used in training teacher and student models.}
  \label{tab:hyperparams}
  \begin{tabular}{p{0.3\linewidth} p{0.3\linewidth}}
    \hline
    \textbf{Parameter} & \textbf{Value} \\
    \hline
    Batch size       & 1024 \\
    Learning rate    & 0.001 \\
    Weight decay     & 0.001 \\
    Epochs           & 50 \\
    Patience         & 8 \\
    Dropout          & 0.3 \\
    Scheduler        & Cosine \\
    Step size        & 10 \\
    Gamma            & 0.5 \\
    \hline
  \end{tabular}
\end{table}

\subsection{Dataset Profiling and Preprocessing}
\label{sec:dataset_char}

The \textsc{TON\_IoT} dataset was generated in a cyber-range integrating edge, fog, and cloud via SDN/NFV orchestration, capturing heterogeneous behaviors across IoT, IIoT, and host systems \cite{moustafa2021new, booij2021heterogeneity}. Its network split contains 21.98 million flow records with 42 features and nine attack categories, including scanning, DoS, DDoS, ransomware, backdoor, injection, Cross-site Scripting, password cracking, and Man-in-the-Middle \cite{tareq2022analysis}. Preprocessing combined a StandardScaler for numeric features and a OneHotEncoder for categorical ones, yielding an expanded input space whose dimensionality and mapping were exported. Partitioning preserved empirical label distributions while minimizing leakage.

\begin{figure}[t]
  \centering
  \includegraphics[width=0.85\columnwidth]{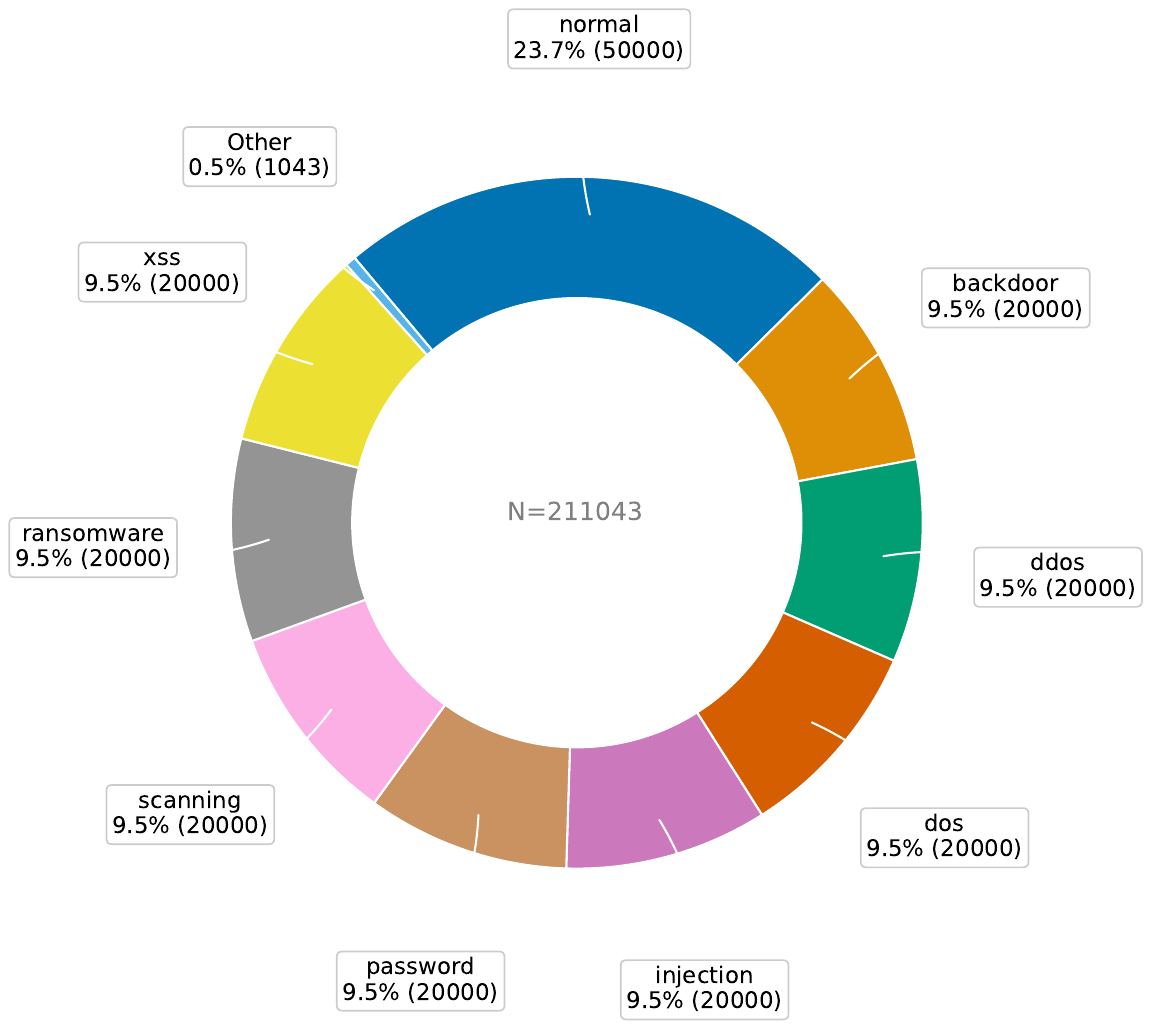}
  \caption{Relative frequencies of attack categories in \textsc{TON\_IoT} network data (donut view with total sample size at center).}
  \label{fig:attack_distribution}
\end{figure}

Figure~\ref{fig:attack_distribution} shows the distribution of attack categories. Volumetric attacks such as DoS and DDoS dominate, while categories such as injection, cross-site scripting, and Man-in-the-Middle are rare, resulting in a heavy-tail imbalance. From a deployment point of view, this imbalance may degrade the recall of minority classes under distribution changes. Mitigation in our workflow includes stratified batching, class-aware metrics, and SHAP-guided pruning combined with knowledge distillation to preserve decision margins for rare attacks under constrained inference.

\begin{figure}[t]
  \centering
  \includegraphics[width=0.85\columnwidth]{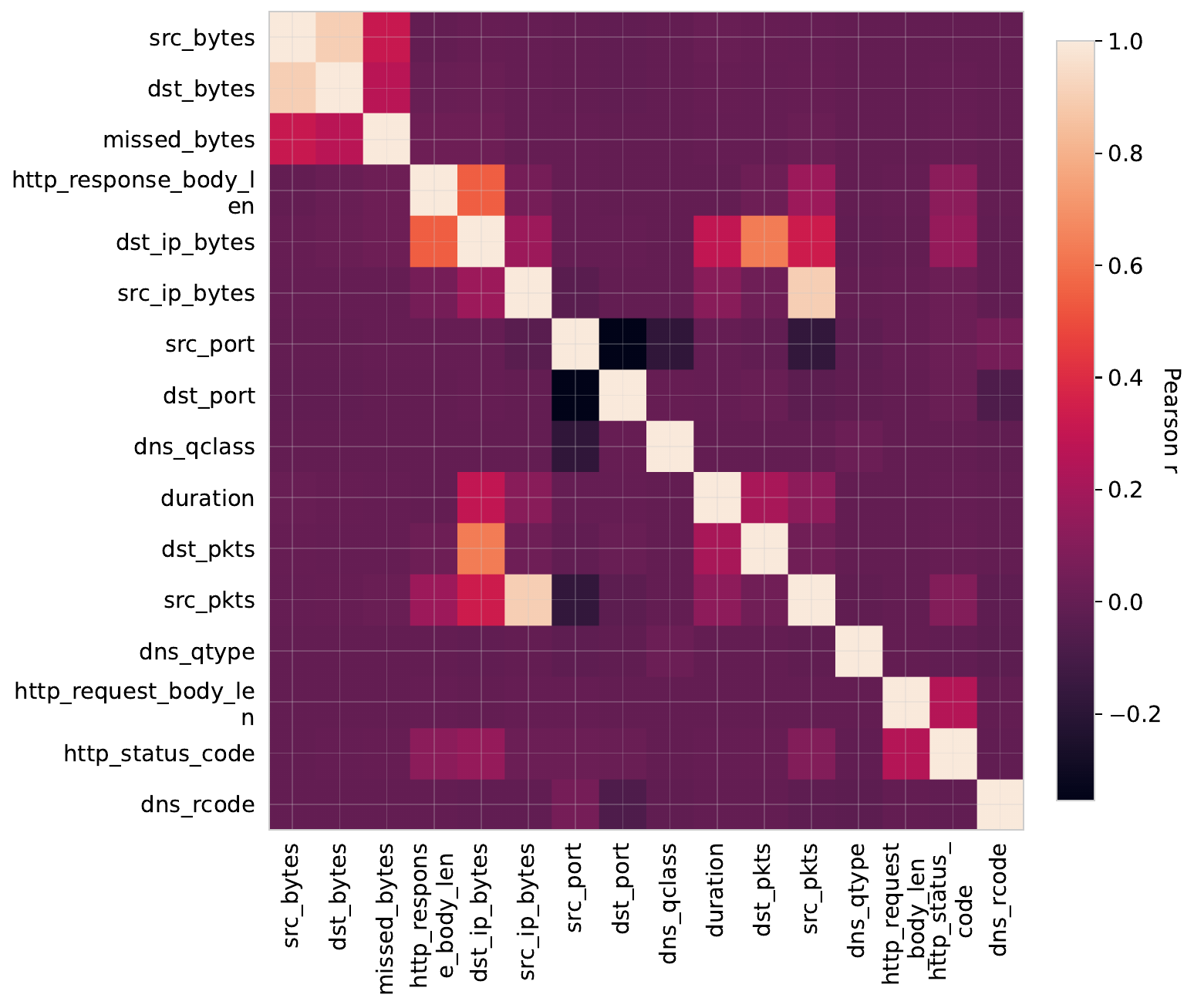}
  \caption{Correlation among top-variance numeric features in the \textsc{TON\_IoT} dataset.}
  \label{fig:correlation_topvar}
\end{figure}

The correlation analysis of high-variance numeric features further reveals redundancy. As shown in Fig.~\ref{fig:correlation_topvar}, traffic volume attributes (\texttt{src\_bytes}, \texttt{dst\_bytes}, \texttt{src\_pkts}, \texttt{dst\_pkts}) exhibit strong dependencies, whereas protocol-level indicators remain weakly correlated. This confirms that SHAP-based feature pruning can remove collinear attributes with little information loss, reducing latency and memory footprint without compromising detection of rare attacks.

\subsection{Experimental Results}

The evaluation integrates explainability-driven feature selection, confusion structure, and runtime benchmarks to assess the student architecture.

\begin{figure}[t]
\centering
\includegraphics[width=0.85\columnwidth]{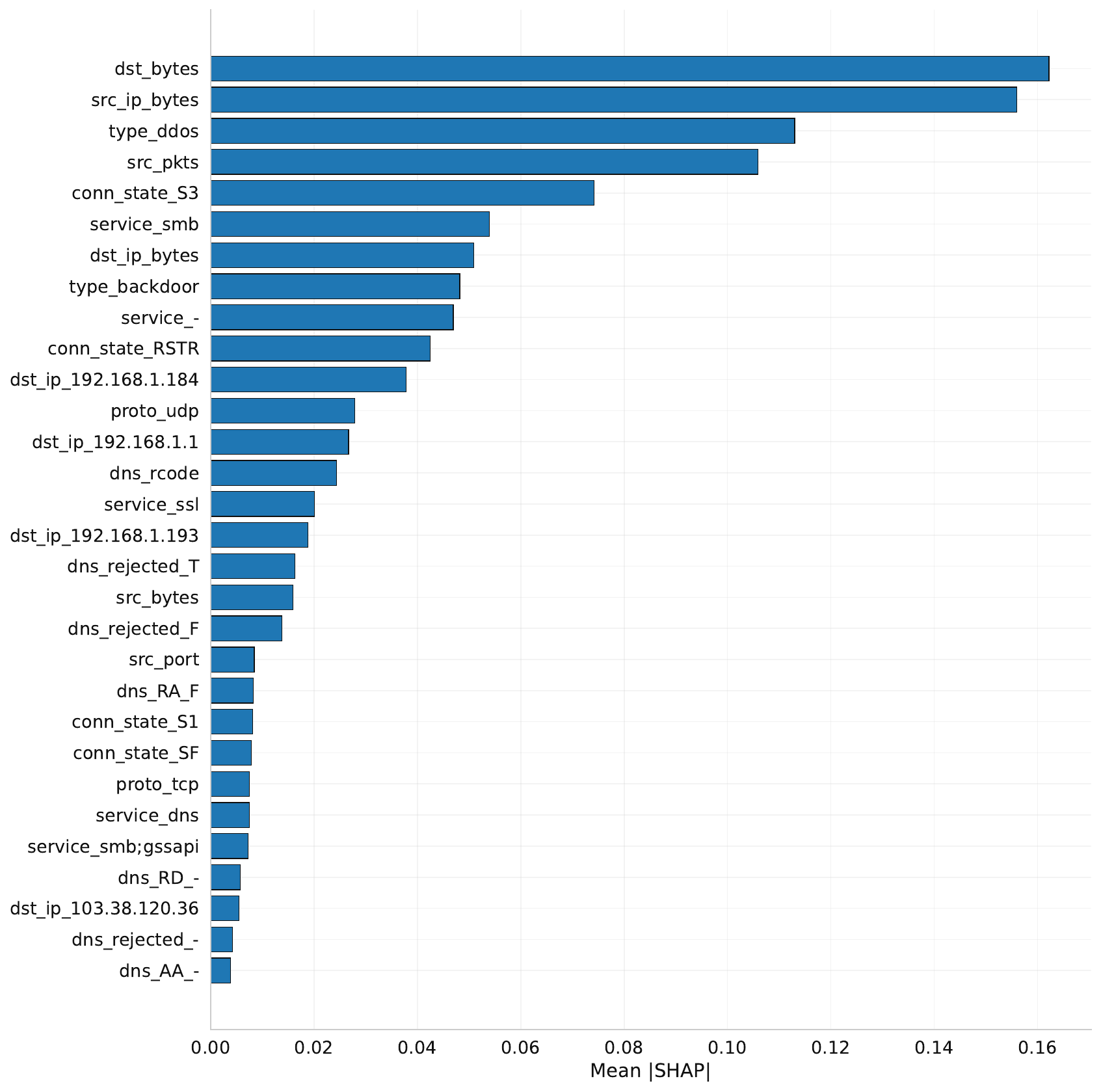}
\caption{Global ranking of discriminative features via mean absolute SHAP.}
\label{fig:shap_global_ranking}
\end{figure}

\begin{figure}[t]
\centering
\includegraphics[width=0.85\columnwidth]{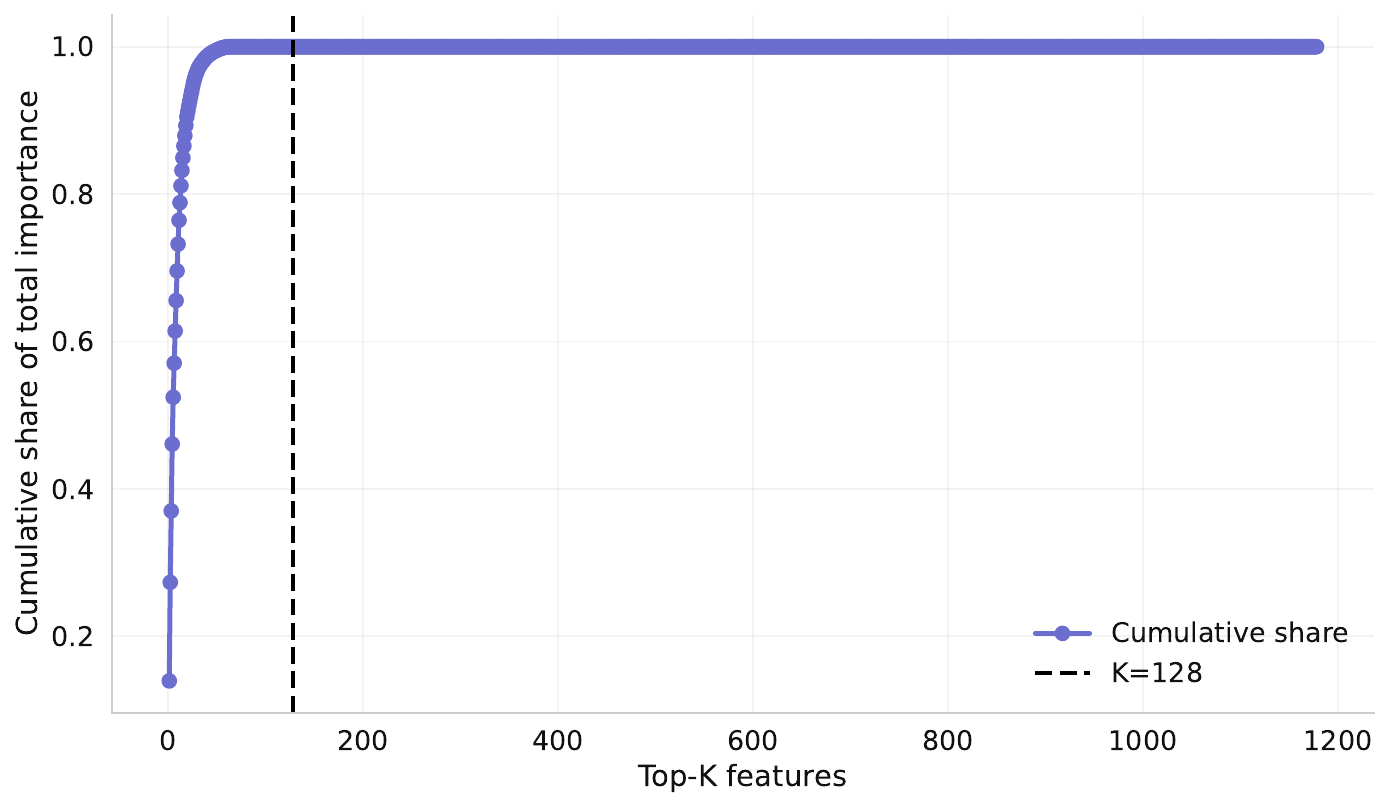}
\caption{Cumulative attribution curve, with cutoff $K=128$ recovering nearly the full importance mass.}
\label{fig:shap_cumulative_curve}
\end{figure}

As shown in Fig.~\ref{fig:shap_global_ranking}, SHAP highlights the traffic volume and connection state attributes as dominant. Moreover, Fig.~\ref{fig:shap_cumulative_curve} indicates that $K=128$ recovers nearly all importance mass. This justifies pruning as a latency-aware strategy, although stability under non-IID traffic and noise remains an open consideration.

\begin{figure}[t]
\centering
\begin{minipage}{0.48\columnwidth}
    \centering
    \includegraphics[width=\linewidth]{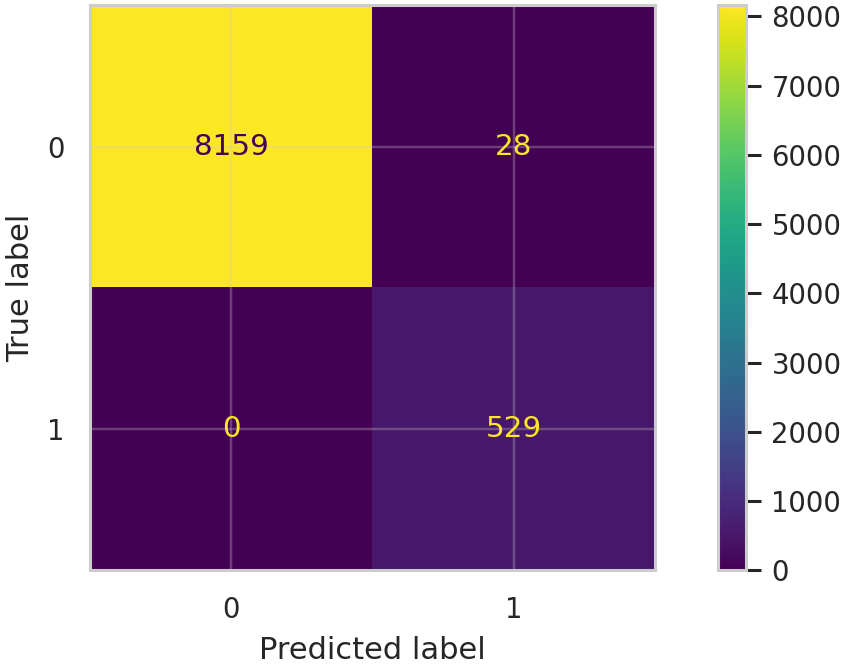}
    \vspace{2pt}
    {\small (a) Student fp32.}
\end{minipage}\hfill
\begin{minipage}{0.48\columnwidth}
    \centering
    \includegraphics[width=\linewidth]{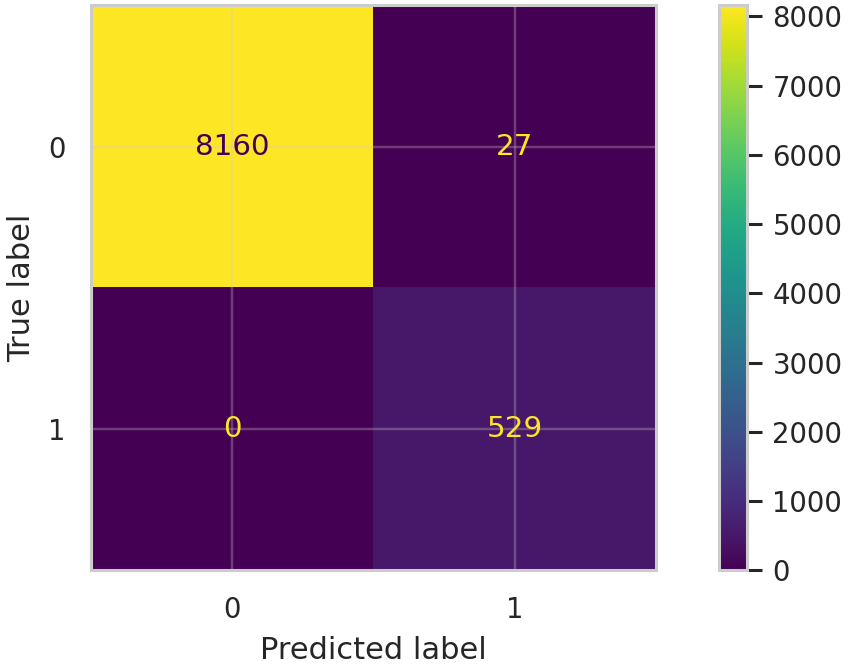}
    \vspace{2pt}
    {\small (b) Student int8.}
\end{minipage}
\caption{Confusion matrices of the distilled student under fp32 and int8 precision.}
\label{fig:student_confmats}
\end{figure}

Fig.~\ref{fig:student_confmats} shows that both fp32 and int8 variants achieve identical recall with no false negatives. Specificity reaches 0.9966 and precision 0.9497 for fp32, with minor improvements for int8. Reported accuracy ($\approx 0.997$) and macro-F1 ($\approx 0.987$) confirm that quantization preserves performance. Errors concentrate in false positives on benign flows, suggesting the need for calibration to mitigate alert fatigue in deployment.

\begin{table*}[ht]
\centering
\caption{Comparison of teacher and student models in terms of accuracy, macro-F1, latency, and compression.}
\label{tab:teacher_student_results}
\begin{tabular}{lcccccccc}
\hline
Model & Accuracy & Macro-F1 & Params & Size (KB) & Mean Lat. (ms) & p95 Lat. (ms) & Speedup & Compression \\
\hline
Teacher (fp32) & 0.9989 & 0.9955 & 769,922 & 3021.53 & -- & -- & -- & -- \\
Student (fp32) & 0.9968 & 0.9863 & 3,042   & 22.29   & 1.29 & 1.62 & 1.00 & 1.00 \\
Student (int8) & 0.9969 & 0.9867 & 3,042   & 22.44   & 1.45 & 2.25 & 0.89 & 0.99 \\
\hline
\end{tabular}
\end{table*}

Table~\ref{tab:teacher_student_results} confirms that the student reduces parameters by nearly $1/250$ while maintaining macro-F1 above 0.986. Both variants preserve perfect recall on attacks, with all residual errors as false positives. The fp32 student achieves mean latency of 1.29 ms and p95 latency of 1.62 ms, enabling real-time inference. The int8 version slightly increases latency but offers additional storage efficiency.

\begin{figure}[t]
\centering
\begin{minipage}{0.9\columnwidth}
    \centering
    \includegraphics[width=\linewidth]{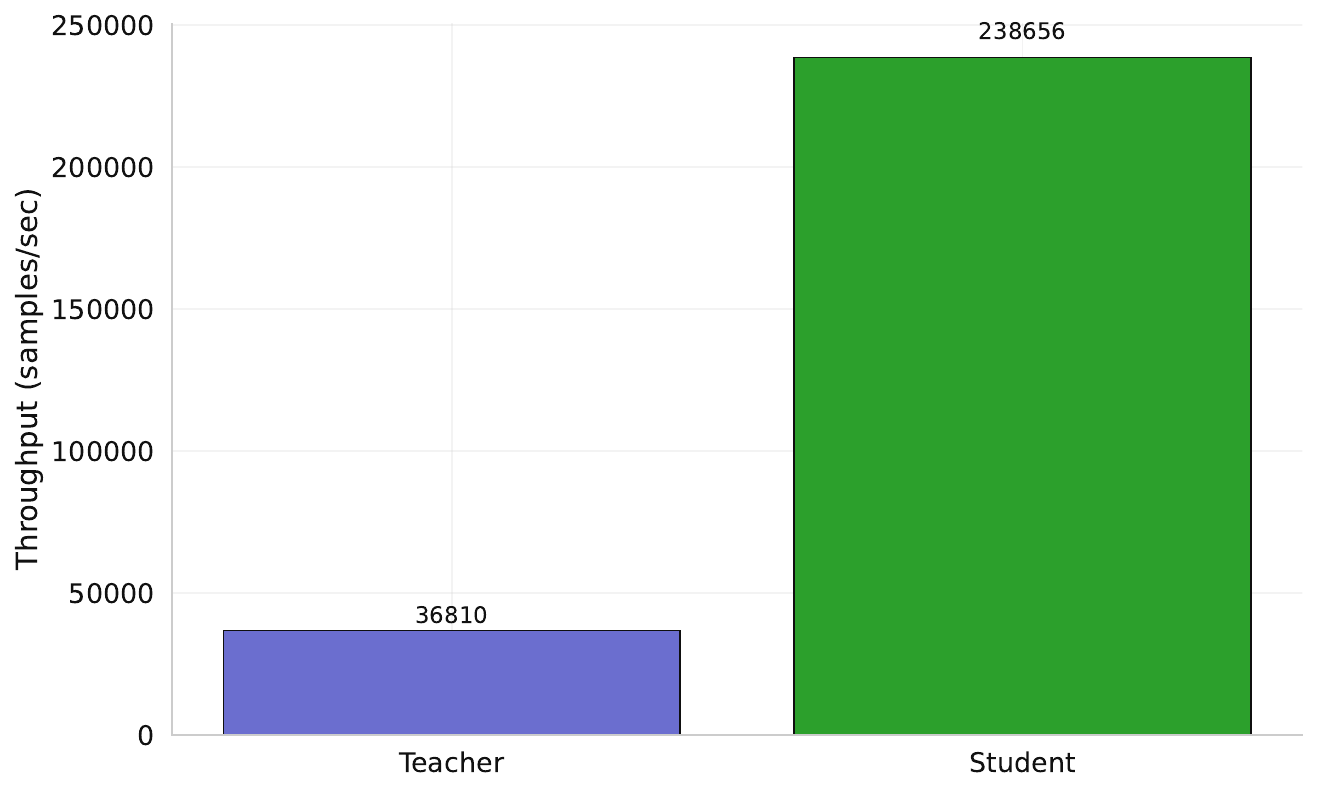}
    \vspace{2pt}
    {\small (a) Throughput over the validation set, in samples/s.}
\end{minipage}

\vspace{6pt}

\begin{minipage}{0.9\columnwidth}
    \centering
    \includegraphics[width=\linewidth]{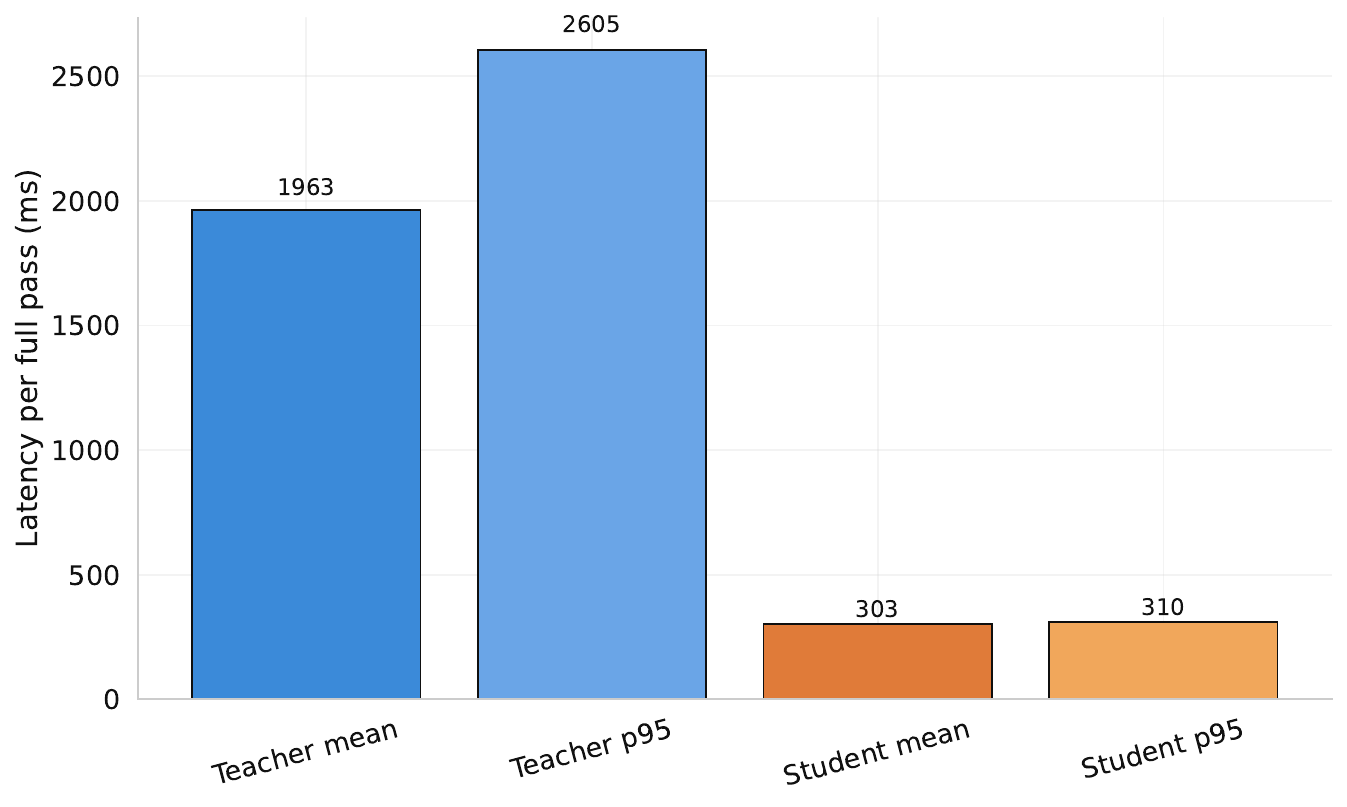}
    \vspace{2pt}
    {\small (b) Latency per validation pass with batch size $1024$, showing mean and p95.}
\end{minipage}

\caption{Runtime comparison between teacher and student models. Panel (a) highlights throughput gains, while panel (b) details latency reductions.}
\label{fig:student_runtime}
\end{figure}

Finally, Fig.~\ref{fig:student_runtime} illustrates runtime benefits. Throughput increases by $\sim 6.5\times$ (from $3.68\times 10^{4}$ to $2.39\times 10^{5}$ samples/s), while mean latency falls from 1963 ms for the teacher to 303 ms for the student. The narrow gap between mean and tail latency indicates stable inference, which is crucial for jitter-sensitive IoT traffic. Nevertheless, the evaluation does not account for heterogeneous deployment conditions, where device-level energy budgets, workload bursts, and non-IID traffic may influence runtime consistency.

\section{Discussion and Research Directions}
\label{sec:discussion}

The near-perfect accuracy achieved by the student model, even after source-aware partitioning and strict preprocessing, indicates that the limitation lies less in the training pipeline than in the dataset structure itself. In the \textsc{TON\_IoT} traces, attack signatures remain highly separable, enabling both teacher and distilled student networks to approximate full memorization. Similar outcomes have been reported in earlier studies, where decision trees or random forests reached comparably high scores on the same benchmark \cite{tareq2022analysis,roy2021lightweight}. 

From a deployment perspective, such results warrant caution. The pipeline prevents leakage, yet the dataset’s restricted adversarial diversity and absence of temporal drift likely inflate accuracy and fail to capture the variability of real traffic. As highlighted in prior work, evolving attack morphologies and non-IID flows typically reduce detection performance in practice \cite{moustafa2021new,booij2021heterogeneity}. Evaluations on idealized corpora may therefore obscure vulnerabilities that emerge in operational IoT scenarios with heterogeneous devices and energy-constrained inference. 

Future research should address these gaps through more challenging evaluation protocols. Cross-dataset validation, for instance, could reveal generalization limits by training on \textsc{TON\_IoT} and testing on X-IIoT-ID or CIC-IDS2021. Robustness may further be examined via controlled label noise, adversarial perturbations, or synthetic datasets that emulate temporal drift. Another open question concerns deployment-level trade-offs: the effect of varying compression ratios on energy–accuracy balance, and the resilience of distilled Kronecker networks to unseen attack variants in federated or highly resource-limited environments. Tackling these questions would move IDS research closer to trustworthy operation in next-generation IoT networks.

\section{Conclusion}
\label{sec:conclusion}
This work presented a lightweight intrusion detection framework for IoT that integrates SHAP-guided feature pruning with knowledge-distilled Kronecker networks, demonstrating that feature-aware pruning and structured compression can drastically reduce model parameters, memory footprint, and inference latency while retaining essential detection capabilities. Experiments confirmed that the student model achieves millisecond-level inference and a footprint below one percent of the teacher, supporting feasibility for deployment in constrained edge devices. However, the results also revealed reduced macro-F1, emphasizing the need for more effective strategies to handle imbalanced traffic and rare attack categories, as well as a deeper understanding of the interaction between pruning depth, distillation temperature, and Kronecker rank in heterogeneous environments. Future directions include adaptive pruning strategies, workload-aware scheduling, privacy-preserving distillation for federated IoT, and cross-dataset stress testing, all aiming to narrow the energy–accuracy gap while ensuring scalability, robustness, and transparency in real-world IoT scenarios.

\bibliographystyle{IEEEtran}
\bibliography{references}

\end{document}